\pdfoutput=1
\documentclass{article} % For LaTeX2e
\usepackage{iclr2015,times}
\usepackage{hyperref}
\usepackage{url}

\usepackage{pbox}
\usepackage{bm}
\usepackage{amsmath}
\usepackage[all]{xy}
\usepackage{graphicx}
\DeclareMathOperator*{\argmax}{arg\,max}
\DeclareMathOperator*{\argmin}{arg\,min}

\usepackage{multirow}
\usepackage{tikz}
\usetikzlibrary{positioning}
\tikzset{%
  every neuron/.style={
    circle,
    draw,
    minimum size=1cm
  },
  neuron missing/.style={
    draw=none, 
    scale=4,
    text height=0cm,
    execute at begin node=\color{black}$\dots$
  },
}
\hypersetup{
colorlinks = true,
citecolor = black,
linkcolor = black,
urlcolor = black,
linkbordercolor = white,
citebordercolor = white,
urlbordercolor = white
}
\title{A Unified Perspective on Multi-Domain and Multi-Task Learning}
\author{
Yongxin Yang \& Timothy M. Hospedales\\
Electronic Engineering and Computer Science\\
Queen Mary, University of London\\
\{yongxin.yang, t.hospedales\}@qmul.ac.uk
}

\iclrfinalcopy % Uncomment for camera-ready version

\iclrconference % Uncomment if submitted as conference paper instead of workshop

\begin{document}

\maketitle

\begin{abstract}
In this paper, we provide a new neural-network based perspective on multi-task learning (MTL) and multi-domain learning (MDL). By introducing the concept of a semantic descriptor, this framework unifies MDL and MTL as well as encompassing various classic and recent MTL/MDL algorithms by interpreting them as different ways of constructing semantic descriptors. Our interpretation provides an alternative pipeline for zero-shot learning (ZSL), where a model for a novel class can be constructed without training data. Moreover, it leads to a new and practically relevant problem setting of zero-shot domain adaptation (ZSDA), which is the analogous to ZSL but for novel domains: A model for an unseen domain can be generated by its semantic descriptor. Experiments across this range of problems demonstrate that our framework outperforms a variety of alternatives.
\end{abstract}

\section{Introduction}

%P1: Introducing & Differentiating MDL and MTL.
Multi-task and multi-domain learning are established strategies to improve learning by sharing knowledge across different but related tasks or domains. Multi-domain learning refers to sharing information about the same problem across different contextual domains, while multi-task learning addresses sharing information about different problems in the same domain. Because the domain/task distinction is sometimes subtle, and some methods proposed for MTL can also address MDL and vice-versa, the two settings are sometimes loosely used interchangeably. However, it is useful to distinguish them clearly: 
%\begin{center}
%\begin{tabular}{ c | l l }
%\hline  & Single Domain & Multiple Domain \\ 
%\hline Single Task & Conventional Supervised Learning & Multi-Domain Learning \\ 
%Multiple Task &  Multi-Task Learning & Multi-Domain-Multi-Task Learning \\ \hline 
%\end{tabular} 
%\end{center}
Domain relates to some covariate, such as the bias implicitly captured in a particular dataset \citep{torralba2011dataset_bias}, or the specific data capture device. For example the Office Dataset \citep{saenko2010adapting} contains three domains related to image source: Amazon, webcam, and DSLR. A multi-domain learning problem can then be posed by training a particular object recogniser across these three domains (same task, different domains). In contrast, a multi-task problem would be to share information across the recognisers for individual object categories (same domain, different tasks). 
%Note that multi-domain learning is less commonly explicitly named in the literature, because many MTL algorithms can also solve it. \textcolor{blue}{
The issue of simultaneously addressing multiple tasks and multiple domains seems to be un-addressed in the literature to our knowledge.

% P2: Our contribution to MDL/MTL.
In this paper, we propose a neural network framework that addresses both multi-domain and multi-task learning, and can perform simultaneous multi-domain multi-task learning.  A key concept in our framework is the idea of a multivariate ``\emph{semantic descriptor}'' for tasks and domains.  Such a descriptor is often available as metadata, and can be exploited to improve information sharing for MTL and MDL.
We show that various classic and recent MTL/MDL methods are special cases of our framework that make particular assumptions about this descriptor: Existing algorithms typically implicitly assume categorical domains/tasks, which is less effective for information sharing when more detailed task/domain metadata is available. For example, the classic ``school dataset" poses a task of predicting students' grades, and is typically interpreted as containing a domain corresponding to each school. However, since each school has three year groups, representing domains by a semantic descriptor tuple (school-id, year-group) is better for information sharing. Our framework  exploits such multi-variate semantic descriptors effectively, while existing MTL/MDL algorithms would struggle to do so, as they implicitly consider tasks/domains to be atomic.

%In this paper, we propose a neural network framework that addresses both multi-domain and multi-task learning. We focus on the concept of a semantic descriptor for tasks and domains, and show how this can be used to improve information sharing for better MTL and MDL. 
%Within this interpretation, we demonstrate that various classic and recent MTL/MDL algorithms are special cases of our framework making particular assumptions about layers, regularisers, and semantic descriptor input. In particular, we find that existing algorithms typically implicitly assume a categorical assumption on domains/tasks, which is less effective for information sharing when more detailed task/domain metadata is available, as is often the case in practice.  For example, the classic "school dataset" is typically assumed to contain a domain corresponding to each school. However, since each school has three year groups, representing domains by the tuple (school-id,year) is better for information sharing. Our framework can exploit such multi-variate semantic descriptors effectively, while existing MTL/MDL algorithms would struggle to do so, as they explicitly consider tasks/domains to be atomic.
%

Going beyond information sharing for known tasks, an exciting related paradigm for task-transfer is ``zero-shot'' learning (ZSL)  \citep{larochelle2008zerodata, lampert2009learning, fu2014embedding}. This setting addresses automatically constructing a test-time classifier  for categories which are unseen at training time. Our neural-network framework provides an alternative pipeline for ZSL. % % Yongxin: while it's not *that* novel compared to zero-data learning (aaai 2008) and DeViSE (nips), and we don't have strong evidence to support "it's better" % %
More interestingly, it leads to the novel problem setting of zero-shot domain adaptation (ZSDA):  Synthesising a model appropriate for a new unseen domain given only its semantic descriptor. For example, suppose we have an audio recogniser trained for a variety of acoustic playback environments, and for a variety of microphone types: Can we synthesise a recogniser for an arbitrary environment-microphone combination? To  our knowledge, this is the first time that zero-shot domain adaptation has been addressed specifically. %Note that ZSDA and zero-shot learning are ``related" but not ``same", despite ZSDA provides an alternative pipeline for zero-shot learning problems.      

%, which are the sources of the images, a multi-domain learning problem can be proposed by training three keyboard-vs-laptop classifiers across these three domains. In contrast, a multi-task problem is usually to decompose a multi-class or multi-label problem with $C$ categories into $C$ one-vs-rest problems and train them simultaneously. Note that multi-domain learning is less commonly used in literature because many MTL algorithms are also able to solve it, e.g., in section 6.3 of \cite{Argyriou2008}, the regression problem for school data is (more) precisely termed a multi-domain problem because there are 139 schools (domains) but one single problem -- to predict the student's exam score. \cite{daume2012gomtl} repeat this experiment using their method (GO-MTL), but they also apply it to MNIST digit recognition by considering 0-9 digits as ten binary one-vs-rest tasks, and we specifically term the later case as multi-task learning for clarity. 

\section{Related Work}

\subsection{Multi-Task Learning}

Multi-Task Learning (MTL) aims to jointly learn a set of tasks by discovering and exploiting task similarity. Various assumptions have been made to achieve this. An early study \citep{Evgeniou2004} assumed a linear model for $i$th task can be written as $w_i := w_0 + v_i$ where $w_0$ can be considered as the \emph{shared knowledge} which benefits all tasks and $v_i$ is the \emph{task-specific knowledge}.% A hierarchical model proposed by \cite{salakhutdinov2011learning} is similar to this motivation, where a tree-structured model for one object is generated by the sum of itself (\emph{task-specific knowledge}) and its parents (\emph{shared knowledge}), i.e., $\bm{w}^{\text{(van)}}:=\bm{w}^{\text{(global)}}+\bm{w}^{\text{(vehicle)}}+\bm{w}^{\text{(van)}}$.  

Another common assumption of MTL is that the predictors (task parameters) lie in a low dimensional subspace \citep{Argyriou2008}. Imposing the (2,1)-norm on the predictor matrix $W$, where each column is a task, results in a low-rank $W$, which implicitly encourages parameter sharing. However, this assumes that all tasks are related, which is likely violated in practice. Forcing predictors to be shared across unrelated tasks can significantly degrade the performance -- a phenomenon called negative transfer \citep{Rosenstein05totransfer}. A task grouping framework is thus proposed by \cite{ICML2011Kang} that partitions all tasks into disjoint groups where each group shares a low dimensional structure. This partially alleviates the unrelated task problem, but misses any fundamental information shared by all tasks, as there is no overlap between the subspaces of each group. 

As a middle ground, the GO-MTL algorithm \citep{daume2012gomtl} allows information to be shared between different groups, by representing the model of each task as a linear combination of latent predictors. Thus the concept of grouping is no longer explicit, but determined by the coefficients of the linear combination. Intuitively, model construction can be thought of as: $W = LS$ where $L$ is the matrix of which each column is a latent predictor (\emph{shared knowledge}), and $S=[s_1,s_2,\ldots,s_M]$ where $s_i$ is a coefficient vector that cues how to construct the model for the $i$th task (\emph{task-specific knowledge}). It is worth noting that this kind of predictor matrix factorisation approach -- $W = LS$ -- can explain several models: \cite{daume2012gomtl} is L1/L2 regularised decomposition, \cite{daume2012flexiblemtl} is linear Gaussian model with IBP prior and an earlier study \citep{Xue2007} assumes $s_i$ are unit vectors generated by a Dirichlet Process (DP). 
 
Most MTL methods in literature assume that each task is an atomic entity indexed by a single categorical variable. Some recent studies \citep{icml2013_romera-paredes13,wimalawarnemultitask} noticed a drawback -- this strategy can not represent a task with more structured metadata, e.g., (school-id, year-group) for school dataset. Thus they replace predictor matrix $W$ with a tensor so that the linear models across more than one categorical variable can be placed in the tensor. Then they follow the line of \cite{Argyriou2008} to impose a variety of regularisations on the mentioned tensor, such as sum of the ranks of the matriciations of the tensors \citep{icml2013_romera-paredes13} and scaled latent trace norm \citep{wimalawarnemultitask}. However, this again suffers from the strong assumption that all tasks are related. 
%\textcolor{red}{Technically producing model in absence of training data is feasible in this framework by some tensor completion techniques but it again suffers the violation of the assumption that all tasks are related. }

\subsection{Multi-Domain Learning}
\paragraph{Domain Adaptation}
There has been extensive work on domain adaptation (DA) \citep{beijborn2012daCV}. A variety of studies have proposed both supervised \citep{saenko2010adapting,duan2012msDA} and unsupervised \citep{gong2012geodesicFlowDA,sun2014objectDetAdapt} methods. As we have mentioned, the typical assumption is that domains are indexed by a single categorical variable: For example a data source such as Amazon/DSLR/Webcam \citep{saenko2010adapting}, a benchmark dataset such as PASCAL/ImageNet/Caltech  \citep{gong2012geodesicFlowDA}, or a modality such as image/video \citep{duan2012msDA}.

Despite the majority of research with the categorical assumption on domains, it has recently been generalised by studies considering domains with a (single) continuous parameter such as time \citep{hoffman2014continuousDA} or viewing angle \citep{qui2012domainAdaptDict}. In this paper, we take an alternative approach to generalising the conventional categorical formulation of domains, and instead investigate information sharing with domains described by a \emph{vector} of discrete parameters.

\paragraph{Multi-Domain Learning}

Multi-Domain Learning (MDL) \citep{DredzeKC10, JoshiDCR12} shares properties of both domain adaptation and multi-task learning. In conventional domain adaptation, there is an explicit pair of source and target domain, and the knowledge transfer is one way Source$\rightarrow$Target. In contrast, MDL encourages knowledge sharing in both directions. %, i.e., Source$\leftrightarrow$Target. 
Although some existing MTL algorithms reviewed in previous section tackle MDL as well, we distinguish them by the key difference during testing time: MDL makes prediction for same problem (binary classification like ``is laptop") across multiple domains (e.g., datasets or camera type), but MTL handles different problems (such as ``is laptop'' versus ``is mouse'').% thus it might be followed a rank step to make prediction for multi-class and multi-label problem.

\subsection{Zero-Shot Learning}

Zero-Shot Learning (ZSL) aims to eliminate the need for training data for a particular task. It has been widely studied in different areas, such as character  \citep{larochelle2008zerodata} and object recognition \citep{lampert2009learning,socher2013zslCrossModal,fu2014embedding}. % and text categorisation \cite{CRRS08, SongSoRo14}. The notion of ZSL varies in different work, for example, it's rephrased as ``Zero-Data Learning'' in \cite{larochelle2008zerodata} and ``Dataless Classification" in \cite{CRRS08}. 
Typically for ZSL, the label space of training and test data are disjoint, so no data has been seen for test-time categories. Instead, test-time classifiers are constructed given some mid-level information. Although diverse in other ways, most existing ZSL methods follow the pipeline in \cite{Palatucci_2009_6459}: $X\rightarrow Z \rightarrow Y$ where $Z$ is some ``semantic descriptor", which refers to attributes \citep{lampert2009learning} or semantic word vectors \citep{socher2013zslCrossModal}. Our work can be considered as an alternative pipeline, which is more similar to \cite{larochelle2008zerodata} and \cite{DeViSE} in the light of the following illustration: $\xymatrix{Z \ar@<0.1pt>@/^0.5pc/[rr] & X \ar[r] & Y}$.

Going beyond conventional ZSL, we generalise the notion of zero-shot learning of tasks to zero-shot learning of domains. In this context, zero-shot means no training data has been seen for the target domain prior to testing. The challenge is to construct a good model for a novel test domain based solely on its semantic descriptor.  %This is the analogous problem for domain-adaptation that zero-shot learning poses for recognition. 
The closest work to our zero-shot domain adaptation setting is \cite{ding2014missingModality}, which addresses the issue of  a missing modality with the help of the partially overlapped modalities that have been previously seen. However they use a single fixed modality pair, rather than synergistically exploiting an arbitrary number of auxiliary domains in a multi-domain way as in our framework. Note that despite the title, \cite{Blitzer_zero-shotdomain} actually considers unsupervised domain adaptation without target domain labels, but \emph{with} target data.

\section{Model}

\subsection{General Framework}
Assume that we have M domains (tasks), and the $i$th domain has $N_i$ instances. We denote the feature vector of the $j$th instance in the $i$th domain (task) and its associated semantic descriptor by the pair $\{\{x^{(i)}_j,z^{(i)}\}_{j=1,2,\cdots,N_i}\}_{i=1,2,\cdots,M}$ and the corresponding label as  $\{\{y^{(i)}_j\}_{j=1,2,\cdots,N_i}\}_{i=1,2,\cdots,M}$. Note that, in multi-domain or multi-task learning, all the instances are effectively associated with a semantic descriptor indicating their domain (task).  Without loss of generality, we propose an objective function that minimises the empirical risk for all domains (tasks),

\begin{equation}
\argmin_{P,Q}\frac{1}{M}\sum_{i=1}^{M}\bigg(\frac{1}{N_i}\sum_{j=1}^{N_i}\mathcal{L}\Big(\hat{y}^{(i)}_j,y^{(i)}_j\Big)\bigg), \quad\text{where}\quad \hat{y}^{(i)}_j = f_{P}(x^{(i)}_j) \cdot g_{Q}(z^{(i)})
\label{eq1}
\end{equation}
% +\Omega_1(P)+\Omega_2(Q)

\begin{figure}
\vspace{-0.25cm}
\centering
\resizebox{13cm}{!}{%
\begin{tikzpicture}[x=1.5cm, y=1.5cm, >=stealth]

\foreach \m/\l [count=\y] in {1,2,3,missing,4}
  \node [every neuron/.try, neuron \m/.try] (input-\m) at (\y,0) {};

\foreach \m [count=\y] in {1,2,3,missing,4}
  \node [every neuron/.try, neuron \m/.try ] (hidden-\m) at (\y,1.5) {};

\foreach \i in {1,...,4}
  \foreach \j in {1,...,4}
    \draw [->] (input-\i) -- (hidden-\j);

\node [align=right] at (0.25,0) {x};
\node [align=right] at (0.25,0.75) {P};
\draw (0.5,-0.5) -- (5.5,-0.5) -- (5.5,0.5) -- (0.5,0.5) -- (0.5,-0.5);
\draw (0.5,1) -- (5.5,1) -- (5.5,2) -- (0.5,2) -- (0.5,1);
\draw (3,2) -- (3,2.5) -- (5.875,2.5);

\pgfmathsetmacro{\PP}{5.75}

\foreach \m/\l [count=\y] in {1,2,3,missing,4}
  \node [every neuron/.try, neuron \m/.try] (input2-\m) at (\PP+\y,0) {};

\foreach \m [count=\y] in {1,2,3,missing,4}
  \node [every neuron/.try, neuron \m/.try ] (hidden2-\m) at (\PP+\y,1.5) {};

\foreach \i in {1,...,4}
  \foreach \j in {1,...,4}
    \draw [->] (input2-\i) -- (hidden2-\j);

\node [align=left] at (\PP+5.75,0) {z};
\node [align=right] at (\PP+5.75,0.75) {Q};

\draw (\PP+0.5,-0.5) -- (\PP+5.5,-0.5) -- (\PP+5.5,0.5) -- (\PP+0.5,0.5) -- (\PP+0.5,-0.5);
\draw (\PP+0.5,1) -- (\PP+5.5,1) -- (\PP+5.5,2) -- (\PP+0.5,2) -- (\PP+0.5,1);
\draw (\PP+3,2) -- (\PP+3,2.5) -- (5.875,2.5);

\draw[thick,->] (5.875,2.5) -- (5.875,2.75);

\node [every neuron] (output1) at (5.875,3.075) {y};

\end{tikzpicture}

}
\vspace{-0.75cm}
\caption{Two-sided Neural Network for Multi-Task/Multi-Domain Learning} \label{fig:model}
\vspace{-0.5cm}
\end{figure}
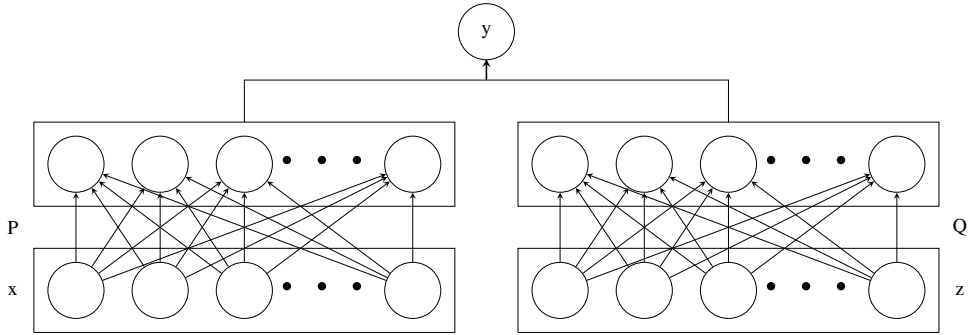

This model can be understood as a two-sided neural network illustrated by Figure~\ref{fig:model}. One can see it contains two learning processes: the left-hand side is representation learning $f_P(\cdot)$, starting with the original feature vector $x$; and the right-hand side is model construction $g_Q(\cdot)$, starting with an associated semantic descriptor $z$. $P$ and $Q$ are the weights to train for each side. % % Yongxin: I label this blue because P/Q is not necessarily to be matrix, at this point, they are still general papameters in NN, e.g., matrics for inner product layer, kernel for convolution layer. Note that we will say they are matrix later. % % Regression or classification predictions for $y$ can be generated by applying data $x$ and task/domain descriptor $z$.
To train $P$ and $Q$, standard back propagation can be performed by the loss $\mathcal{L}(\cdot)$ calculated between ground truth $y$ and the prediction $\hat{y}$.

%\begin{figure}
%\centering
%\includegraphics[width=0.7\linewidth]{./model}
%\caption{NAME NEEDED HERE}
%\label{fig:model}
%\end{figure}

%\vspace{-0.9cm}
With this neural network interpretation, the two sides can be arbitrarily complex but we find that one inner product layer for each is sufficient to unify some existing MDL/MTL algorithms and demonstrate the efficacy of the approach. In this case, $P$ is a D-by-K matrix and $Q$ is a B-by-K matrix, where K is the number of units in the middle layer; D and B is the length of feature vector $x$ and semantic descriptor $z$ respectively. The prediction is then based on $(x^{(i)}_j P) (z^{(i)} Q)'$. %{$\hat{y}^{(i)}_j = x^{(i)}_j P (z^{(i)} Q)'$.} 

\subsection{Unification of Existing Algorithms}

We next demonstrate how a variety of existing algorithms\footnote{RMTL: \textbf{R}egularized \textbf{M}ulti--\textbf{T}ask \textbf{L}earning, FEDA: \textbf{F}rustratingly \textbf{E}asy \textbf{D}omain \textbf{A}daptation, MTFL: \textbf{M}ulti--\textbf{T}ask \textbf{F}eature \textbf{L}earning and GO-MTL: \textbf{G}rouping and \textbf{O}verlap for \textbf{M}ulti--\textbf{T}ask \textbf{L}earning} are special cases of our general framework. For clarity we show this in an MDL/MTL setting  with $M=3$ domains/tasks. Observe that RMTL \citep{Evgeniou2004}, FEDA\footnote{a is an (M+1)-dimensional row vector with all ones, e.g., $a=[1,1,1,1]$ when M=3, b is a D-by-D identity matrix, and $\otimes$ denotes Kronecker product. $w_0$, $w_1$, $w_2$, $w_3$ and $\underline{0}$ in $Q'$ are D-dimensional column vectors.} \citep{iii2007frustratingly}, MTFL \citep{Argyriou2008} and GO-MTL \citep{daume2012gomtl} each assume specific settings of $Z$, $P$ and $Q'$ as in Table~\ref{tab:modeldemo}. 

\begin{table}[h]
\vspace{-0.5cm}
\caption{A Unifying Review of Some Existing MTL/MDL Algorithms}
\begin{center}
\scalebox{0.9}{
\begin{tabular}{c c c c c c}
\hline  & $Z$ & $P$ & Norm on $P$ & $Q'$ & Norm on $Q'$ \\ \hline
RMTL & $\begin{bmatrix}1 & 0 & 0 & 1 \\0 & 1 & 0 & 1 \\0 & 0 & 1 & 1  \end{bmatrix}$ & Identity & None & $\begin{bmatrix}| & | & | & | \\v_1 & v_2 & v_3 & w_0\\| & | & | & |  \end{bmatrix}$ & None \\  
FEDA$^2$ & $\begin{bmatrix}1 & 0 & 0 & 1 \\0 & 1 & 0 & 1 \\0 & 0 & 1 & 1  \end{bmatrix}$ & $a\otimes b$ & None & $\begin{bmatrix}\underline{0} & \underline{0} & \underline{0} & w_0 \\ w_1 & \underline{0} & \underline{0} & \underline{0}\\\underline{0} & w_2 & \underline{0}& \underline{0}\\\underline{0} & \underline{0} & w_3 &\underline{0} \end{bmatrix}$ & None \\
MTFL & $\begin{bmatrix}1 & 0 & 0 \\0 & 1 & 0 \\0 & 0 & 1  \end{bmatrix}$ & Identity & None & $W$ & (2, 1)-Norm \\ 
GO-MTL & $\begin{bmatrix}1 & 0 & 0 \\0 & 1 & 0 \\0 & 0 & 1 \end{bmatrix}$ & $L$ & Frobenius & $S$ & Entry-wise $\ell_1$ \\ 
\hline
\end{tabular}}
\end{center}
\label{tab:modeldemo}
\vspace{-0.33cm}
\end{table}
% \footnote{a is a $M+1$ dimensional row vector with all ones, i.e., $a=[1,1,1,1]$, b is a D-by-D identity matrix, and $\otimes]$ is Kronecker product. $w_0$ and $0$ in $Q'$ is D-dimensional vector.}
% $\begin{bmatrix}1 & 1 & 0 & 0 \\1 & 0 & 1 & 0 \\1 & 0 & 0 & 1  \end{bmatrix}$

The notion used is kept same with the original paper, e.g., $P$ here is analogous to $L$ in \cite{daume2012gomtl}. Each row of the matrices  in the second ($Z$) column is the corresponding domain's semantic descriptor in different methods. These methods are implicitly assuming a single categorical domain/task index: with 1-of-N encoding as semantic descriptor (sometimes with a constant term). % % Yongxin: should we call the all ones column 'constant term' or 'intercept term' or 'bias term' % %

We argue that more structured domain/task-metadata is often available, and with our framework it can be directly exploited to improve information sharing compared to simple categorical indices. For example, suppose two categorical variables (A,B) describe a domain, and each of them has two states (1,2), then four distinct domains can be encoded by $Z$ in a distributed fashion (Table~\ref{tab:codingdemo} left) in contrast to the 1-of-N form used by traditional multi-task learning methods (Table~\ref{tab:codingdemo} right). The ability to exploit more structured domain/task descriptors $Z$ where available, improves information sharing  compared to existing MTL/MDL methods. In our experiments, we will demonstrate  examples of problems with multivariate domain/task metadata, and its efficacy to improve learning. 
%\vspace{-0.15cm}
\begin{table}
\vspace{-0.85cm}
\caption{Illustration for Distributed Coding and 1-of-N Coding}
\vspace{-0.5cm}
\begin{center}
\begin{equation*}
{\small
\begin{bmatrix}
 & \text{A-1} & \text{A-2} & \text{B-1} & \text{B-2} \\
\text{Domain-1} & 1 & 0 & 1 & 0 \\
\text{Domain-2} & 1 & 0 & 0 & 1 \\
\text{Domain-3} & 0 & 1 & 1 & 0 \\
\text{Domain-4} & 0 & 1 & 0 & 1
\end{bmatrix}  
\begin{bmatrix}
 & \text{A-1-B-1} & \text{A-1-B-2} & \text{A-2-B-1} & \text{A-2-B-2} \\
\text{Domain-1} & 1 & 0 & 0 & 0 \\
\text{Domain-2} & 0 & 1 & 0 & 0 \\
\text{Domain-3} & 0 & 0 & 1 & 0 \\
\text{Domain-4} & 0 & 0 & 0 & 1
\end{bmatrix}
}
\end{equation*}
\vspace{-0.75cm}
\label{tab:codingdemo}
\end{center}
\end{table}

%\vspace{-0.25cm}
%We argue that such vector domain-metadata is in fact widely available and can be used to boost  multi-task and multi-domain learning. 

%\vspace{-0.25cm}
\subsection{Learning Settings}

\paragraph{Multi-domain multi-task (MDMT)} 
Existing frameworks have focused on either MDL or MTL settings but not considered both together. Our interpretation provides a simple means to exploit them both simultaneously for better information sharing when multiple tasks in multiple domains are available. If $z^{(d)}$ and $z^{(t)}$ are the domain and task descriptors respectively, then MDMT learning can be performed by simply concatenating the descriptors $[z^{(d)}, z^{(t)}]$ corresponding to the domain and task of each individual instance during learning. 
%\textcolor{red}{\emph{TMH: at test time how does this relate to  statement made elsewhere that task descriptor not present at test}}
\vspace{-0.1cm}

\paragraph{Zero-shot learning (ZSL)} 
As mentioned, the dominant zero-shot (task) learning pipeline is $X\rightarrow Z \rightarrow Y$.  At train time, the $X\to Z$ mapping is learned by classifier/regressor, where $Z$ is a task descriptor, such as a binary attribute vector  \citep{lampert2009learning,fu2014embedding}, or a continuous word-vector describing the task name \citep{socher2013zslCrossModal,fu2014embedding}. At testing time, the ``prototype'' semantic vector for a novel class $z$ is presented, and zero-shot recognition is performed by matching the $X\to Z$ estimate and prototype $z$, e.g., by nearest neighbour \citep{fu2014embedding}.

In our framework, ZSL is achieved by presenting each novel semantic vector $z^*_j$ (each testing category is indexed by $j$) in turn along with novel category instances $x^*$. Zero-shot recognition then is given by: $j^*=\argmax_{j} f_P(x^*)\cdot f_Q(z^*_j)$. 
\vspace{-0.1cm}

\paragraph{Zero-shot domain adaptation (ZSDA)} The zero-shot domain adaptation task can also be addressed by our framework. With a distributed rather than 1-of-N encoded domain descriptor, only a subset of domains is necessary to effectively learn $Q$. Thus a model suitable for data from a \emph{novel held-out domain} can be constructed by applying its semantic descriptor $z^*$ along with data $x^*$. %, and predicting \textcolor{blue}{$y=f_P(x^*)\cdot f_Q(z^*)$}.
\vspace{-0.1cm}

\section{Experiments}
\vspace{-0.1cm}

We demonstrate our framework on five experimental settings: MDL, ZSDA, MTL, ZSL and MDMT. %For this evaluation, we use two classic multi-task benchmarks as well as two pattern recognition tasks: audio classification and image-based animal recognition. 

\textbf{Implementation:} We implement the model with the help of Caffe framework \citep{jia2014caffe}. Though we don't place regularisation terms on $P$ or $Q$, a non-linear function $\sigma(x)=max(0,x)$ (i.e., ReLU activation function) is placed to encourage sparse models $g_Q(z^{(i)})=\sigma(z^{(i)}Q)$. The choice of loss function for regression and classification is Euclidean loss and Hinge loss respectively. Preliminary experiments show $K=\frac{D}{\log(D)}$  leads to satisfactory solutions for all datasets.

%TMH: Does this mean that the current predictive statement (just above Sec 3.2) is inaccurate for MTL, only applies to MDL?

\textbf{MTL/MDL Baselines:} We compare the proposed method with a single task learning baseline -- linear or logistic regression with $\ell_2$ regularisation (LR), and four multi-task learning methods: (i) RMTL \citep{Evgeniou2004}, (ii) FEDA \citep{iii2007frustratingly}, (iii) MTFL \citep{Argyriou2008} and (iv) GO-MTL \citep{daume2012gomtl}. Note that these methods are re-implemented within the proposed framework. We have verified our implementations with the original ones and found that the performance difference is not significant. Baseline methods use traditional 1-of-N encoding, while we use a distributed descriptor encoding based on metadata for each problem.  

\textbf{Zero-Shot Domain Adaptation:} We follow the MDL setting to learn $P$ and $Q$ except that one domain is held out each time. We construct test-time models for held out domains using their semantic descriptor. We evaluate against two baselines: (i) Blind-transfer (LR): learning a single linear/logistic regression model on aggregated data from all seen domains. To ensure fair comparison, distributed semantic descriptors are concatenated with the feature vectors for baselines, i.e., they are included as a plain feature. (ii) Tensor-completion (TC): we use a tensor $W \in \mathcal{R}^{D, p_1, p_2, \cdots, p_N}$ to store all the linear models trained by SVM where $N$ is the number of categorical variables and $p_i$ is the number of states in the $i$th categorical variable ($p_1 + p_2 + \cdots + p_N = B$ in our context and $p_1 * p_2 * \cdots * p_N = M$ if there is always a domain for each of possible combinations). ZSDA can be formalised by setting the model parameters for the held-out domain to missing values, and recovering them by a low-rank tensor completion algorithm \citep{KressSV:2014}. This low-rank strategy corresponds to our implementation of  \cite{icml2013_romera-paredes13}.

%\noindent\textbf{Zero-Shot Learning:} We use a classic ZSL benchmark where the mid-level information naturally serves as semantic descriptor. For example, in Animals with Attributes (AwA) dataset \citep{lampert2009learning}, the the semantic descriptor for cat is its binary attribute vector.
%Note that the testing set is disjoint with the training set across all domains and it is same for both MDL and ZSDA.

\subsection{School Dataset - MDL and ZSDA}

%We first evaluate two classic benchmarks, because these have been extensively evaluated by most prior studies.

\noindent\textbf{Data}\quad
This classic dataset\footnote{Available at http://multilevel.ioe.ac.uk/intro/datasets.html} collects exam grades of 15,362 students from 139 schools. Given the 23 features\footnote{The original dataset has 26 features, but 3  that indicate student year group are used in semantic descriptors.}, a regression problem is to predict each student's exam grade. There are 139 schools and three year groups. School IDs and year groups naturally form multivariate domains. Note that 64 of 139 schools have the data of students for all three year groups, and we also choose the school of which each year group has more than 50 students so that each domain has sufficient training data. Finally there are $23\times3=69$ distinct domains given these two categorical variables.

%\textbf{Forest Fires}\quad
%Forest Fire \cite{cortez2007} is a regression task to predict the burned area of forest fires using meteorological and other features. We choose ``12 months" and ``workday versus weekend" as two variables to construct %$12\cdot2=24$ domains. %As preprocessing, we impose a smooth log function $y:=\log(y+1)$ on target variable -- burned area as it is extremely biased. 

\textbf{Settings and Results}\quad
For MDL we learn all domains together, and for ZSDA we use a leave-one-domain-out strategy, constructing the test-time model based on the held-out domain's descriptor with $P$ and $Q$ learned from the training domains. In each case the training/test split is 50\%/50\%. Note that the test sets for MDL and ZSDA are the same. The results in Table~\ref{tab:classic} are averages over the test set for all domains (MDL), and averages over the held-out domain performance when holding out each of the 69 domains in turn (ZSDA). Our method outperforms the alternatives in each case.

%\begin{table}[h]
%  \caption{Classic Datasets: School and Forest Fire (RMSE)}
%\begin{center} 
%
%  \begin{tabular}{l l l l l}
%    \hline
%    & \multicolumn{2}{c}{School} & \multicolumn{2}{c}{Forest Fire}\\\hline
%        & MDL & ZSDA & MDL & ZSDA \\
%LR&9.51&10.35&1.64&1.51\\
%SVR&9.89&10.95&1.55&1.51\\
%GO-MTL&10.00&-&1.64&-\\
%RMTL&00.00&-&0.00&-\\
%MTFL&00.00&-&0.00&-\\
%TC&-&00.00&-&0.00\\
%Ours&\textbf{9.37}&\textbf{10.19}&\textbf{1.45}&\textbf{1.48}\\\hline
%  \end{tabular}
%  \label{tab:classic}
%\end{center}
%\end{table}

\begin{table}[h]
  \caption{School Dataset (RMSE)}
\begin{center}
  \begin{tabular}{l r r r r r r r}
    \hline
     & LR & RMTL & FEDA & MTFL & GO-MTL & TC & Ours \\\hline
    MDL & 9.51 & 9.46 & 10.75 & 10.22 & 10.00 & - & \textbf{9.37} \\
    ZSDA & 10.35 & - & - & - & - & 12.41 & \textbf{10.19} \\
    \hline
    \end{tabular}
  \label{tab:classic}
\end{center}
\end{table}

\subsection{Audio Recognition - MDL and ZSDA}

Audio analysis tasks are affected by a variety of covariates, notably the playback device /  environment (e.g., studio recording versus live concert hall), and the listening device (e.g., smartphone versus professional microphone). Directly applying a model trained in one condition/domain to another will result in poor performance. Moreover, as the covariates/domains are combinatorial: (i) models cannot be trained for all situations, and (ii) even applying conventional domain adaptation is not scalable. Zero-shot domain adaptation has potential to address this, because a model could be calibrated on the fly for a given environment.

\textbf{Data}\quad We investigate recognition in a complex set of noise domains: covering both acoustic environment and microphone type. We consider a music-speech discrimination task introduced by \cite{Tzanetakis:2002zr}, which includes 64 music and speech tracks. Two categorical variables are \emph{smartphone microphone} and \emph{live concert hall} environment, and each of them has two states: on or off. Then the four domains are generated as: (i) Original (ii) Live Recording (LR) (iii) Smartphone Recording (SR) and (iv) smartphone in a live hall (LRSR). The noises are synthesised by Audio Degradation Toolbox \citep{matthias2013a}.

\textbf{Settings and Results}\quad
We use MFCC % \citep{Ellis05-rastamat}
 to extract  audio features and K-means to build a K = 64 bag-of-words representation. We split the data 50\%/50\% for training and test and keep test sets same for MDL and ZSDA. 
%There is no overlap training and testing across all domains, so if a track is chosen for training, it won't appear in testing for any domain. 
The results in Table~\ref{tab:audio2} break down the results by each domain and overall (MDL), and each domain when held-out (ZSDA). In each case our method is best or joint-best due to better exploiting the semantic descriptor (recall that it does not have any additional information; for fairness the descriptor is also given to the other methods as a regular feature). The only exception is the least practical case of ZSDA recognition in a noise free environment given prior training only in noisy environments. The ZSDA result here generally demonstrates that models can be synthesised to deal effectively with new multivariate domains / covariate combinations without needing to exhaustively see data and explicitly train models for all, as would be conventionally required.
\vspace{-0.25cm}
\begin{table}[h]
\caption{Audio Recognition: Music versus Speech (Error Rate)}\label{tab:audio2}
\begin{center}
  \begin{tabular}{ l l  r  r  r  r r }
    \hline
& & Origin & LR & SR & LRSR & Avg  \\ \hline
\multirow{5}{*}{\rotatebox[origin=c]{90}{MDL}}& LR&\textbf{3.13}&18.75&6.25&17.19&11.33\\
& RMTL&6.25&18.75&6.25&17.19&12.11\\
& FEDA&7.81&18.75&9.38&18.75&13.67\\
& MTFL&6.25&21.88&9.38&\textbf{14.06}&12.89\\
& GO-MTL&\textbf{3.13}&\textbf{17.19}&6.25&18.75&11.33\\
& Ours&\textbf{3.13}&\textbf{17.19}&\textbf{4.69}&\textbf{14.06}&\textbf{9.77}\\
\hline
\multirow{3}{*}{\rotatebox[origin=c]{90}{ZSDA}}&LR&\textbf{32.81}&28.13&14.06&23.44&24.61\\
& TC&46.88&21.88&26.56&59.38&38.67\\
& Ours&35.94&\textbf{9.38}&\textbf{12.50}&\textbf{18.75}&\textbf{19.14}\\\hline
  \end{tabular}
\end{center}
\end{table}
%\vspace{-0.75cm}
\subsection{Animal with Attributes - MTL and ZSL}

Animal with Attributes \citep{lampert2009learning} includes images from 50 animal categories, each with an 85-dimensional binary attribute vector. The attributes, such as ``black", ``furry", ``stripes", describe an animal semantically, and provide a unique mapping from a combination of attributes to an animal. The original setting of ZSL with AwA is to split the 50 animals into 40 for training and hold out 10 for testing. 
%ZSL typically involves two steps: (i) predict the attributes for a novel category image (multi-label classification), (ii) compare the predicted attributes with the ground truth of each animal's attributes. The assumption is that the mapping from image feature to attributes is sufficiently general, so the learned mapping from training is applicable to novel animal categories in the test set. 
%The motivation of ZSL is to reduce the workload of labelling, because it's much easier to associate an animal with an attribute vector rather than assign the label of such animal to a number of images. 
We evaluate this condition to investigate: (i) if multi-task learning of attributes and classes improves over the STL approaches typically  taken when analysing AwA, (ii) if it helps to use the attributes as an MTL semantic task descriptor against the traditional setting of MTL where semantic descriptor is a 1-of-N unit vector indexing tasks. For MTL training on AwA, we decompose the multi-class problem with $C$ categories to $C$ one-vs-rest binary classification tasks. Note that in this case the semantic descriptor reveals the label, so it is not given during testing. We run all one-vs-rest classifiers on each instance and rank the scores to produce the label. 

\textbf{Multi-Task Learning}\quad We use the recently released DeCAF feature \citep{Donahue_ICML2014} for AwA. For MTL, we pick five animals from the training set with moderately overlapped attributes, and use the first half of the images for training then test on the rest. The results in Table~\ref{tab:awa} show limited improvement by existing MTL approaches over the standard STL. However, our attribute-descriptor approach to encoding tasks for MTL improves the accuracy by about $2\%$ over STL.

\begin{table}[h]
\caption{AwA: MTL Multi-Class Accuracy}\label{tab:awa}
\begin{center} 
\begin{tabular}{l r r r r r r }
\hline
& antelope&killer whale&otter&walrus&blue whale&Avg\\\hline
LR&92.31&87.08&89.26&75.60&82.44&85.34\\
RMTL&86.08&71.22&80.99&61.90&\textbf{96.18}&79.28\\
FEDA&92.31&83.39&88.15&79.17&89.31&86.47\\
MTFL&92.67&85.61&90.36&79.76&87.02&87.09\\
GO-MTL&91.21&84.87&89.81&\textbf{80.36}&84.73&86.20\\
Ours&\textbf{93.41}&\textbf{91.51}&\textbf{94.21}&79.76&79.39&\textbf{87.66}\\\hline
\end{tabular}
\end{center}
\end{table}

\textbf{Zero-Shot Learning}\quad For ZSL, we adopt the training/testing split in \cite{lampert2009learning}. The blind-transfer baseline is not meaningful because there are different binary classification problems, and aggregating does not lead to anything. Also, tensor-completion is not practical because of its exponential space ($D*2^{85}$) against $D*40$ observations. Our method achieves 43.79\% multi-class accuracy, compared to 41.03\% from direct-attribute prediction (DAP) approach \citep{lampert2009learning} using DeCAF features.
%, where a set of independently trained SVMs (single task learning) is used to predict the attributes and the estimated attribute vector is matched to the prototype (ground truth attribute vector). 
A recent result using DeCAF feature is 44.20\% in \cite{MurphyECCV14}, but this uses additional higher order attribute correlation information. Given that we did not design a solution for AwA specifically, or exploit this higher order correlation cue, the result is encouraging.

\subsection{Restaurant \& Consumer Dataset - MDMT}

The restaurant \& Consumer Dataset, introduced by \cite{vargas2011effects} contains 1161 customer-to-restaurant scoring records, where each record has 43 features and three scores: food, service and overall. We build a multi-domain multi-task problem as follows: (i) a domain refers to a restaurant, (ii) a task is a regression problem to predict one of the three scores given an instance and (iii) an instance is a 43-dimensional feature vector based on customer's and restaurant's profile. The 1161 records cover 130 restaurants but most of them just have few scores, so we just pick 8 most frequently scored ones, and we split training and test sets equally. The semantic descriptor is constructed by concatenating 8-bit domain and 3-bit task indicator. Conventional MTL interpretations of this dataset consider $8\times3=24$ atomic tasks. Thus the task overlap across domain or domain overlap across task is ignored. Results in Table \ref*{tab:rc} shows that our approach outperforms this traditional MTL setting by better representing it as a distributed MDMT problem. 
%\vspace{-0.25cm}
\begin{table}[h]
  \caption{Restaurant \& Consumer Dataset (RMSE)}
\begin{center}
  \begin{tabular}{r r r r r r}
    \hline
      LR  & RMTL & FEDA & MTFL & GO-MTL & Ours \\
     2.32 & 1.23 & 1.17 & 1.13 & 1.06 & \textbf{0.78} \\
    \hline
    \end{tabular}
  \label{tab:rc}
\end{center}
\end{table}
\vspace{-0.50cm}
\section{Conclusion}

In this paper we proposed a unified framework for multi-domain and multi-task learning. The core concept is a semantic descriptor for tasks or domains. This can be used to unify and improve on a variety of existing multi-task learning algorithms. Moreover it naturally extends the use of a single categorical variable to index domains/tasks to the multivariate case, which enables better information sharing where additional metadata is available. Beyond multi task/domain learning, it enables the novel task of zero-shot domain adaptation and provides an alternative pipeline for zero-shot learning. 

%TMH: We should say something like this in case we get a reviewer from the perspective of MTL/MDL is only a problem when you didn't learn invariant features to start with.
Neural networks have also been used to address MTL/MDL by learning shared invariant features \citep{Donahue_ICML2014}. Our contribution is complementary to this (as demonstrated e.g., with AwA) and the approaches are straightforward to combine by placing more complex structure on left-hand side $f_P(\cdot)$. Our future directions are: (i) The current semantic descriptor is formed by discrete variables. We want to extend this to continuous and periodic variable like the pose, brightness and time. (ii) We assume the semantic descriptor (task/domain) is always observed, an improvement for dealing with a missing descriptor is also of interest.   

\noindent\textbf{Acknowledgements} We gratefully acknowledge the support of NVIDIA Corporation for the donation of the GPUs used for this research. 

%\vspace{-0.5cm}

\bibliography{../../../../Library}
\bibliographystyle{iclr2015}

\end{document}